\documentclass[3p,times,procedia]{elsarticle}
\usepackage{ecrc}
\usepackage[bookmarks=false]{hyperref}
    \hypersetup{colorlinks,
      linkcolor=blue,
      citecolor=blue,
      urlcolor=blue}

\volume{00}

\firstpage{1}

\journalname{Procedia Computer Science}

\runauth{Author name}

\jid{procs}

\usepackage{amssymb}

\usepackage[figuresright]{rotating}
\usepackage{cleveref}
\usepackage{url}

\usepackage{float}
\usepackage{graphicx}
\usepackage{caption}
\begin{document}
\begin{frontmatter}



\dochead{5th International Conference on Industry 4.0 and Smart Manufacturing}

\title{Potentials of the Metaverse for Robotized Applications\\ in Industry 4.0 and Industry 5.0}


\author{Eric Guiffo Kaigom} 

\address{Frankfurt University of Applied Sciences,	Nibelungenplatz 1, 60439, Frankfurt, Germany}

\begin{abstract}
As a digital environment of interconnected virtual ecosystems driven by measured and synthesized data, the Metaverse has so far been mostly considered  from its gaming perspective that closely aligns with online edutainment. Although it is still in its infancy and more research as well as standardization efforts remain to be done, the Metaverse could provide considerable advantages for smart  robotized applications in the industry. Workflow efficiency, collective decision enrichment even for executives, as well as  a natural, resilient, and sustainable robotized assistance for the workforce are potential advantages.  Hence, the Metaverse could consolidate the connection between Industry 4.0 and Industry 5.0. 
This paper identifies and puts forward  potential advantages of the Metaverse for robotized applications and highlights how these advantages support goals pursued by the \mbox{Industry 4.0 and Industry 5.0 visions.}
\end{abstract}

\cortext[cor1]{Eric Guiffo Kaigom. Tel.: +49-69-1533-3951}
\end{frontmatter}

\email{kaigom@fb2.fra-uas.de}



\section{Introduction}
\label{intro}

The Industry 4.0 vision targets a holistic and transformative incorporation of information and communication to improve responsiveness, production efficiency,  in addition to  self-controlled mass personalization of products and services~\cite{Sch:uh:20}. Decentralized and reconfigurable robotized automation \cite{Morgan:uh:20} along with intelligent production and servicing~\cite{Sanchez:uh:20} are key pillars of this smart automation. At the heart of this level of automation are often interconnected and interoperating  robots. They are endowed with on-board Artificial Intelligence$\slash$Machine Learning(AI$\slash$ML) capabilities to synchronize and orchestrate themselves, predict next actions, and anticipate events while  making skillful, robust,  and accelerated decisions. 
Insights gained from measured data  help enterprises tailor and scale their  productivity and servicing capabilities to  adapt to varying demands and elevate customer and partner experiences. Toward this end,  the Industry 5.0 vision  focuses on the individual, sustainable, and resilient empowerment of the workforce to streamline automation efficiency. Workers are augmented  with natural and uplifting capabilities that harness personal skills in human-automation-collaboration \cite{Noor-A-Rahim}. They are exposed to decent and  engaging work conditions, including AI$\slash$ML-driven  inference and generation of competitive facts \cite{Xian:20}, that not only build upon a critical perception but also reasoning on top of an efficient and dynamic machine-to-machine (M2M) \cite{Xu:uh:20,Noor-A-Rahim} and human-to-machine communication. A  comfortable,  responsive, differentiated, and fulfilling accommodation of customer preferences along with an inclusive proximity \cite {Nicoletti} to capture their subtle needs and offer them an  enhanced quality of \mbox{value are expected \cite{Nicoletti,Xu:uh:20,Xian:20,Noor-A-Rahim}.} 

  \begin{figure*}[t!]
	\includegraphics[width=0.87\columnwidth]{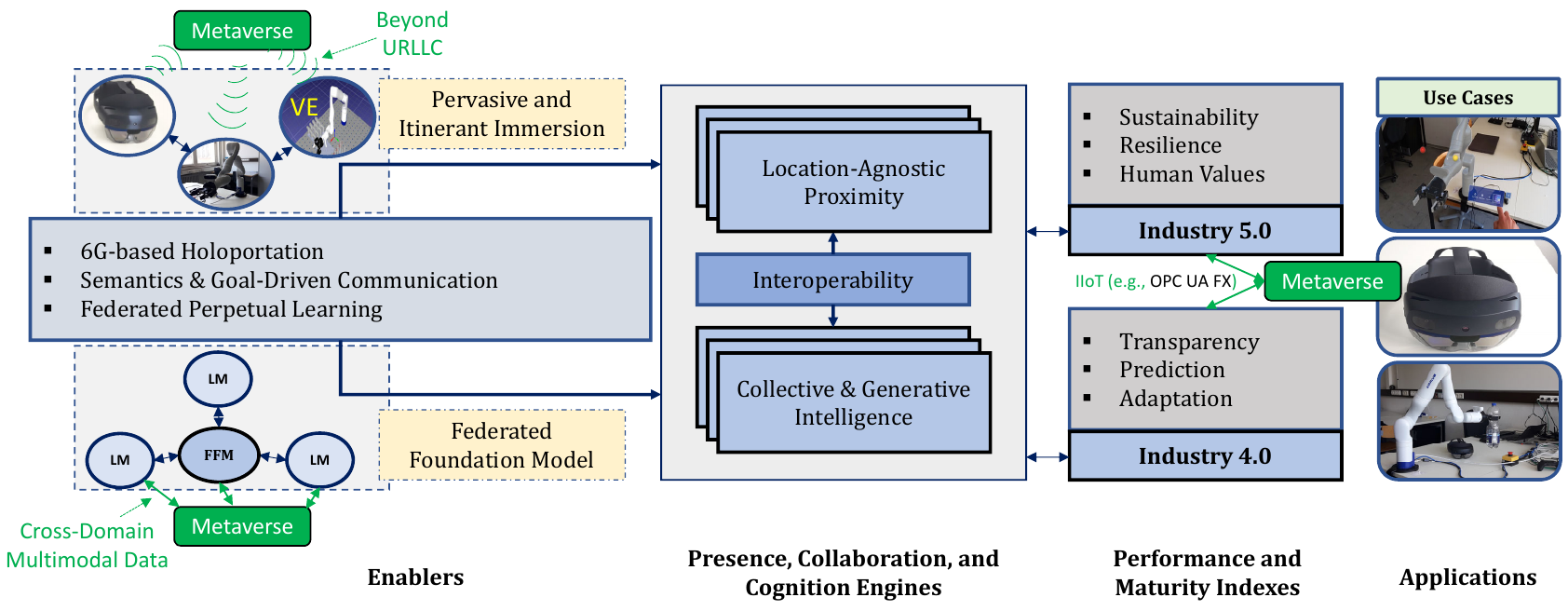}
	\centering
	\caption{The Metaverse contextualized w.r.t some of its enabler emerging technologies as well as core functionalities and performance indexes.}
	\label{basicidea2}
\end{figure*}

 The integration of cognitive skills of humans   and robots to  enhance not only situational awareness but also adaptation through multi-perception and reasoning improves the accommodation of  uncertainties, thereby enhancing the effectiveness of robotized applications that follow the Industry 4.0 vision \cite{Zheng:uh:20}. Take for instance the case of a carpenter using  cobots, i.e., collaborative robots, to industrially construct personalized furniture. The carpenter can even interpret outputs of  overlooked modalities beyond audio and video signals. Forces and torques, for example, can act as a modal means of communication between the carpenter and the cobot. Then, the carpenter correlates these modalities with a balanced understanding of provided recommendations from the analysis of big data (e.g., in-house orders, customer relationship management, states of processes and assets, partner demands and offers,  resources trading, trends, political news, etc.) to make non-nominal anticipatory decisions. Unlike standard automation, the carpenter in the loop can develop an ergonomics-related empathy for customers  involved in the individualized  robotics-driven  design of a chair  whereas the executive department of the fabric can have ethical and moral considerations for customer preferences \cite {Pupa:uh:20}. This human-centered and inclusive servicing  occurs in parallel to a  kinesthetic guidance of the  cobot to naturally and flexibly guide large and heavy payloads to  desired poses. The goal is to responsively reconfigure the workcell and  adapt to high-mix low-volume demands with negligible metabolic costs and zero programming efforts~\cite{wang:uh:20}. In this case, the  compliant behavior of the  cobot contributes to an engaging work experience for  the carpenter, as pursued by the Industry 5.0 vision. The resulting physical human-robot-collaboration (pHRC) fosters  a resilient industry propelled by the capability of the cobot to accommodate   vacancies of co-workers of \mbox{the carpenter in  collaborative tasks \cite{Anil:Kumar}. } However, connecting  robotized applications in Industry 4.0 and Industry 5.0 has received little attention thus far.
 
\begin{figure*}[t!]
 	\includegraphics[width=0.93\columnwidth]{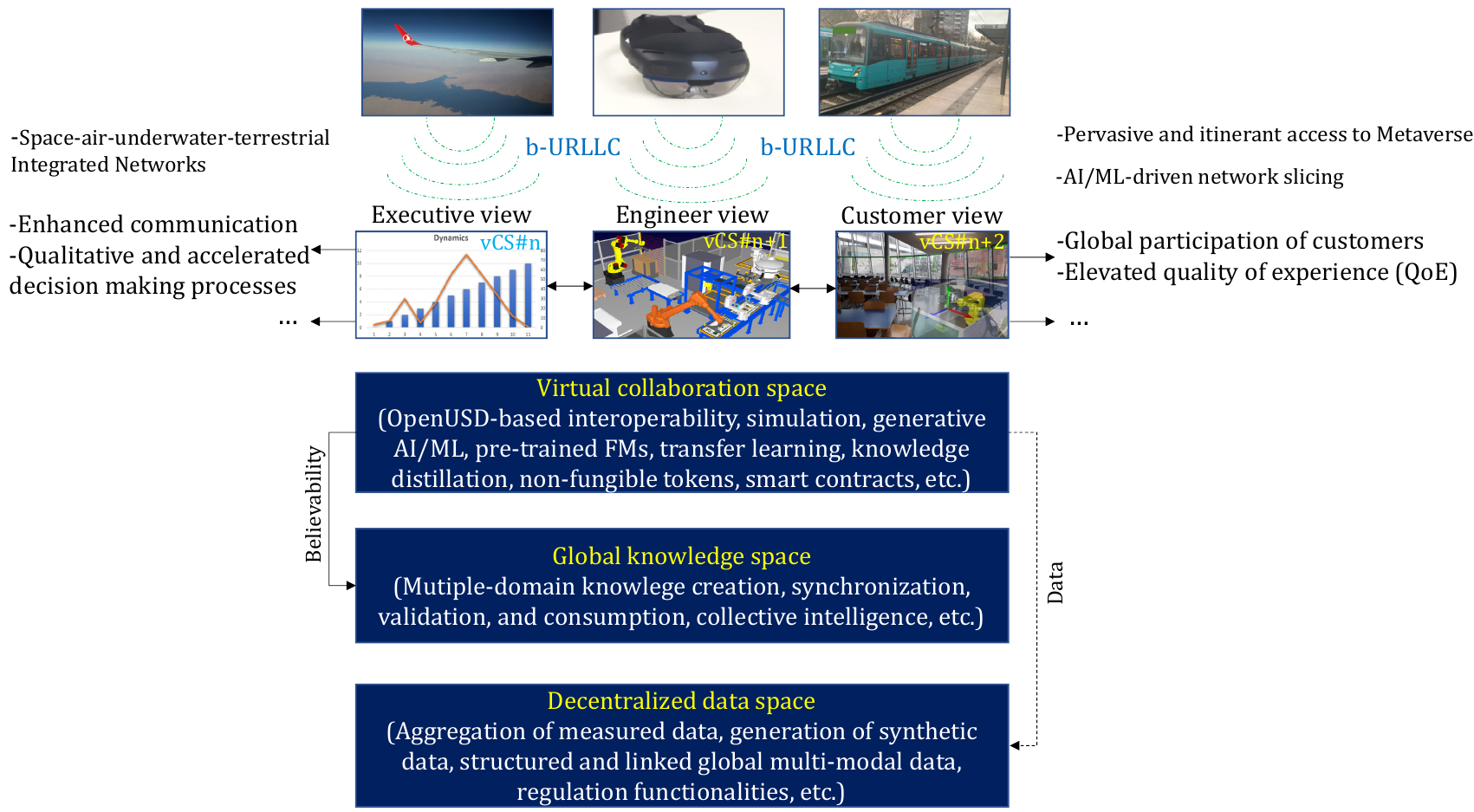}
 	\centering
 	\caption{A three-layer abstraction of the Metaverse.}
 	\label{metaverseconcept2}
 \end{figure*}

 One approach to fill in this gap is to share three mechanisms shown in \cref{basicidea2}. First, a location-agnostic proximity for   collaboration between geographically distributed participants (see \cref{metaverseconcept2}). 
 Secondly, generative \cite {Sohn}   and collective \cite{LiT} intelligence. Generative intelligence interpolates data in the latent space to create synthetic data \cite {Struski}.  Collective intelligence fuses and completes  distributed knowledge graphs to palliate the incompleteness of single one \cite{Zhang}, maximize  shared experiences despite model compression through distillation \cite{Zhang}, and  improve  accuracy  of decision  models in virtual collaboration spaces (vCS) \cite{Yang:Hu}. Models can also map  a task objective to a distribution of configurations \cite {Whitman:Hu} to spawn  possible solutions for practice. Finally,  transfer learning \cite {Sohn} and resource-friendly distillation \cite{Zhang,Yang:Hu}  accelerate  resource-efficient specializations of solutions from multiple foundation models to downstream tasks \cite {Sohn}
 \mbox{in Industry 4.0 and Industry 5.0.} 
 
 
 The Metaverse can be viewed as an ecosystem of interconnected vCS in which this undertaking is developed. vCS are fed with  data and knowledge provided by two functionalities:   generative AI$\slash$ML and collective intelligence (see \cref{metaverseconcept2}). Furthermore, vCS are populated with  digital twins (DTs) of mirrored physical assets and applications in which these assets evolved along with embodied avatars that project humans onto vCS. One advantage of vCS is the flexibility, velocity, and density with which synthetic data is globally generated to simulate and experiment DTs and get insights. Emerging properties are transformed into structured information. The representation of structured facts as knowledge graphs can be learned to infer new knowledge  and inform  decision making, as shown in \cref{metaverseconcept2}. Outcomes of  decisions are shared in the Metaverse or used to meet goals in physical robotized applications (see \cref{metaverseconcept0}). 
 For that,   control policies  learned in the Metaverse using multiple data modalities globally aggregated are transferred to address challenges in reality \cite{Chen}. Transfer learning is the third functionality that helps apply  solutions developed in the Metaverse to challenges in practice.  The three functionalities (i.e., generative AI$\slash$ML, collective intelligence, and transfer learning) contribute to consolidate a bidirectional transition between robotized applications that follows  Industry 4.0 and Industry 5.0. A way to accomplish this consolidation is to develop an  arbitration of the level of shared control or autonomy of the robot (see \cref{metaverseconcept0})  \cite{Selvaggio}. The  arbitration signal is  adapted  to customized goals or inferred intentions \cite{Selvaggio}, ranging from production efficiency (Industry 4.0) driven by autonomous robots to satisfying and resilient work conditions and \mbox{experiences enabled by an intuitive and accessible pHRC (Industry 5.0).}

 \section{Contributions}
 This work identifies and discusses potential advantages the intelligent Metaverse can provide to establish a connection between robotics-driven applications that build upon the Industrial 4.0 and Industry 5.0 visions. It puts forward the concept of shared autonomy$\slash$control  presented in \cite{Selvaggio} and employs it in this work in the context of the Metaverse as a potential approach to achieve the connection illustrated in \cref{metaverseconcept0}.  The paper points out  functionalities in \cref{metaverseconcept2,metaverseconcept0}, which can be leveraged by the Metavere, to impact the \mbox{connection via the shared autonomy$\slash$control by}
 \begin{itemize}
 	\item enriching robotized applications with synthetic data to accelerate workflow and provide insights.
 	\item streamlining decision making in robotics through collective and diversified knowledge creation and usage.
 	\item accelerating decisions as well as  their resource-saving transfer and intuitive use in real robotized applications.
 \end{itemize}
 
 \section{Related Work}
 Although the Metaverse is gaining popularity in  online gaming and edutainment, representatives of global companies such as Nokia and Microsoft expect the biggest impact of the Metaverse in the industry~\cite{Kshetri:20}. The growing interest of the industry in  the Metaverse is related to the large-scale and automatized access to  measured and synthetic data,  flexibility in terms of reachability of vCS from different platforms, virtual immersion and experimentation, scalability in simulation,  as well as the intrinsic safety it offers to industrial partners for the entire product lifecycle management \cite {Chen}. This interest can be illustrated by the cooperation between BMW and NVIDIA. It has led to the Omniverse platform that offers open concepts like  Universal Scene Description (OpenUSD) for interoperability between and integration of scenes in vCS. It thereby streamlines collaboration processes in robotized car assembly and reveals technical and economical implications of chances in workcells  across distributed factories~\cite{urlbmw}. Similarly, Siemens and NVIDIA are  collaborating toward the Industrial Metaverse to bring robotized automation to the next edge \cite{urlsiemens}. Advances in these and similar partnerships have been achieved by integrating emerging technologies. These include generative, collective, and transferable AI$\slash$ML along with virtual$\slash$augmented reality (VR$\slash$AR) for spatial immersion in industrial processes, data analytics to uncover  competitive trends, together with holistic and high-fidelity simulation of DTs to control their physical siblings like factories for  \mbox{cross-domain "what-if" analysis and better-informed decision making.} 
 
  \begin{figure*}[t!]
 	\includegraphics[width=0.8\columnwidth]{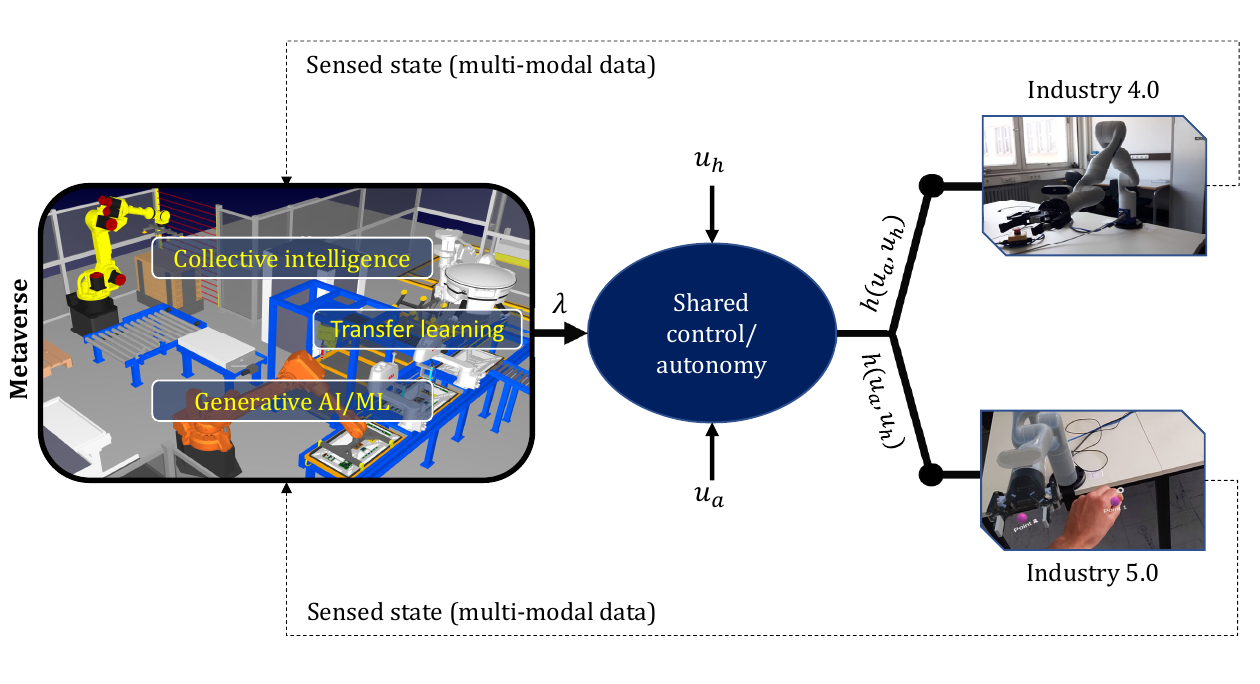}
 	\centering
 	\caption{The Metaverse harnesses shared control$\slash$autonomy, as introduced in \cite{Selvaggio}, to mediate between  robot autonomy ($h(u_a,u_h)=u_a$, see \cref{equation}) for process efficiency and  human supervision of the robot ($h(u_a,u_h)=u_h$, see \cref{equation}) for comfortable, sustainable, and resilient work conditions. }
 	\label{metaverseconcept0}
 \end{figure*}

 For this purpose, high-performance data management for vCS, such as multi-users$\slash$services'  live data synchronization based upon  the Nucleus database of NVIDIA, as well as  accelerated data processing driven  by graphical and tensor processing units paired with high-end rendering that leverages real-time ray-traced lighting   are   harnessed together at the forefront of the development and usage of robotized applications~\cite{urlnvidia,urlissac,Trinh:22}. Simulation allows to quickly model and test possible demands from prospective robotics-relevant markets under almost arbitrary  conditions before financial resources and hardware are committed, thus mitigating multiple risks, such as downtime and logistics issues  \cite{Kshetri:20}. 
 Since the Metaverse hosts DTs from distinct domains, it is a hub for heterogeneous multi-modal data. Such data contribute to diversification and robustification while   pre-training large scale foundation models (FMs) which are then specialized to downstream tasks with comparatively less efforts than starting to learn features from scratch. Enterprises can quickly transfer broad FMs from the \mbox{Metaverse to downstream tasks to circumvent their data and skill limitations.}

\begin{figure*}[t!]
 	\includegraphics[width=0.95\columnwidth]{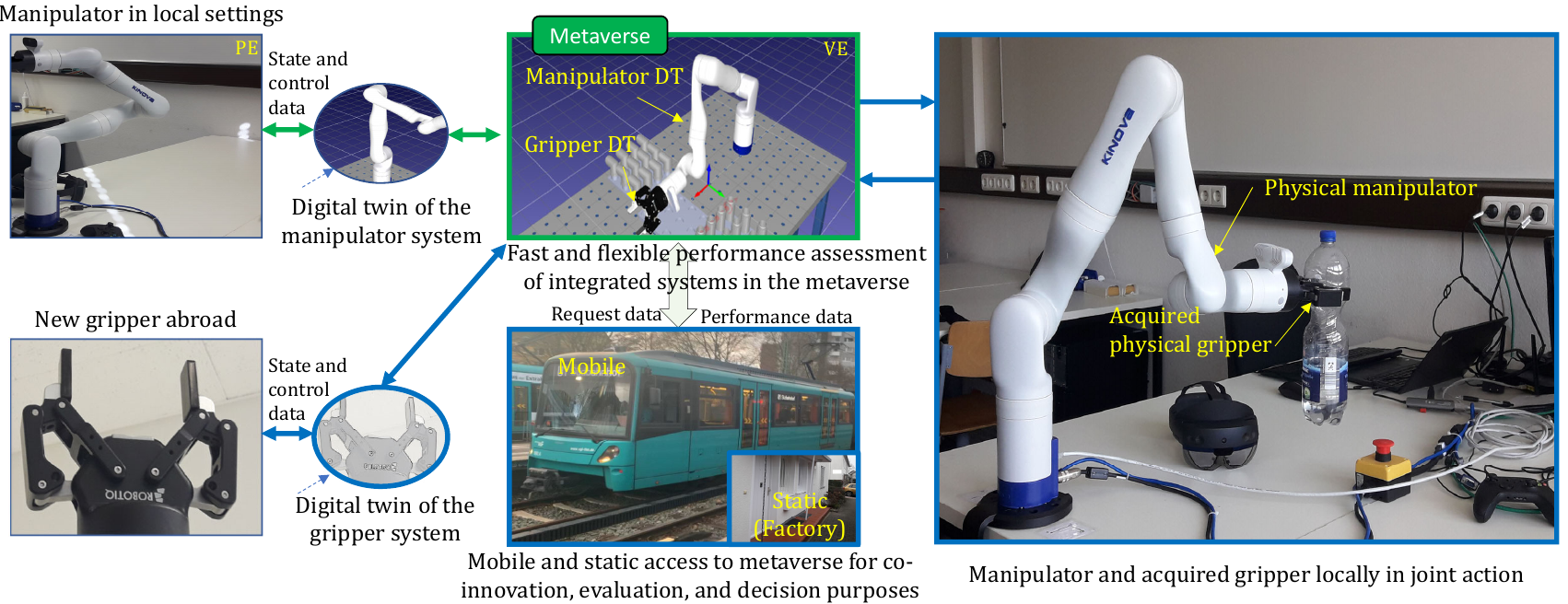}
 	\centering
 	\caption{The  Metaverse as a virtual testbed for  scalable and cost-effective evaluations of  robotics technologies, regardless of the current  mobility.}
 	\label{metaverseconcept}
 \end{figure*}

 Immersive technologies, such as information-rich VR/AR, have been suggested and widely adopted to achieve visibility and transparency in the third and fourth maturity stages associated with Industry 4.0, as outlined by the German Academy of Science and Engineering  (Acatech)~\cite{Sch:uh:20}. In  robotized applications that pursue  objectives of the Industry 5.0 concept, VR$\slash$AR technologies improve   problem-solving competencies (e.g., analytical and critical thinking, etc.) \cite {Gug:erçin:20}, support interpersonal skill transfers (e.g., digital remote assistance, etc.) \cite {Alhloul:20}, and foster  risk awareness (by using e.g. visual  cues) for safety purposes and ergonomically comfortable work in HRC \cite{Cha:cko20}. Efficiency in  predictive maintenance for robotized production lines can benefit from  VR$\slash$AR  because of the quick Metaverse-mediated and thus widely accessible immersion in otherwise invisible robot states (e.g., the safety-critical effective mass at the end-effector) using overlaid visual cues and the  instantaneous proximity to remote robotized applications using upcoming beyond ultra-reliable low-latency wireless communication (b-URLLC) like 6G~\cite{weiliuwerft}. Whereas the complete virtualization and partial augmentation of local real scenes are usual nowadays, a pervasive and itinerant usage of such VR$\slash$AR-solutions even under high mobility (e.g., in a high-speed train as shown in \cref{metaverseconcept}) remains, however, an open challenge with diverse industrial and societal implications. Upon availability, DTs of a manipulator and a prospective gripper can be combined in a vCS of the Metaverse to assess the performance of the compound system,  as shown in the middle upper part of \cref{metaverseconcept}. Potential implications (e.g., impacts of the geometry and pneumatic force range of the gripper on bin picking processes with overlapped items) of the gripper for the velocity of the feeding system in the shop floor can be detected and competitive solutions can be devised in the vCS  long before the gripper is bought. This is done using VR-glasses and DTs while keeping the physical robot in productive usage. Technical and operational questions about how to deal with future co-workers as well as remote  customers willing to co-design   products (e.g., in 3D robotized printing) are still open. These aspects are however crucial for executives in many companies. One reason for this is that future  generations alpha and  Z of workers and customers advocate the necessity of providing interconnected  inclusive workplaces \cite{Jan:ssen:20}, \mbox{in conjunction with market places with personalized experiences and sustainable practices \cite{Are:ola20}.} 
 
Despite the potentials and opportunities of the Metaverse emphasized thus far,  a bidirectional robotics-related transition  between Industry 4.0 and Industry 5.0 is still in its infancy. Such a transition is expected to contribute to identify and consolidate the symbiotic interdependence between both visions and help re-orient  research, development, and transfer activities toward their combination to revamp an efficient industry of values. Whereas an autonomous  behavior of robots that builds upon M2M  communication for process efficiency \cite {ElBanhawy} is a prominent operating mode in Industry 4.0 \cite{Ghodsian}, manual robot guidance based upon e.g. the gravity and payload compensation mode is an approach to cope with musculoskeletal stress in Industry 5.0 \cite{leao}. In-between, the arbitration to control the autonomy level of the robot between full autonomy and full human supervision (e.g., manual guidance) can be a control authority signal $h(u_h,u_a)$ designed as
\begin{equation}\label{equation}
h(u_h,u_a) = \lambda u_h + (1-\lambda)u_a, 
\end{equation}
as introduced in \cite{Selvaggio}. Herein, $u_a$ is the autonomous control input of the robot and $u_h$ is the human intervention. In shared control, a human customizes $\lambda$ \cite{Selvaggio} using the Metaverse functionalities. In shared autonomy, $\lambda$ is automatically tuned \cite{Selvaggio} in the Metaverse using  sensed data and inferences about workers and robots. These data can be enriched with synthetic data, structured, and processed through e.g. knowledge federation to make non-nominal suitable decisions condensed in the variable $\lambda$  as output (see \cref{metaverseconcept0}). For $\lambda=0$, the robot is fully autonomous and there is no intervention of human. Within  the scope of this paper,  full manual guidance can be enforced for $\lambda=1$, which reflects a full control of the human over the  compliant robot for kinesthetic task completion. Therefore, the level of intervention of a human can be expressed by the selection of the Metaverse-mediated scalar $\lambda$ (see \cref{metaverseconcept0}). This intervention together with its natural customization to reach out to a broader range of companies and even novice workers can be influenced by the quality of information gained from the AI$\slash$ML-enabled Metaverse at the different levels of generative, collective, and transferable functionalities instilled  \mbox{in  recent developments of the Metaverse (see \cref{metaverseconcept2}).} 

 \section{Metaverse and its potential}

\subsection{Metaverse and enabler technologies}
The Metaverse spans  vCS  interoperating with each other using  open information model like OpenUSD (see \cref{metaverseconcept2}). vCS host  virtualized environments while some of them persistently reflect physical entities through tightly coupled data-driven synchronization. Entities include factories, workers, and underlying processes as well as robot manipulators and their  workspaces. These entities evolve and interact with each other in vCS as DTs or embodied avatars. Most vCS support custom and individualized  OPC UA$\slash$MQTT-based and VR$\slash$AR-oriented interfaces for optimization-dedicated machine-to-DT communication and people immersion. b-URLLC is expected to provide ultra-high reliability  and ultra-low latency for stable control loops and high-fidelity   tactile transmissions (e.g., less force distortion and corrupted multi-modal sensations) \cite{Marshall} in the vCS-based architecture in \cref{metaverseconcept0}. b-URLLC could elevate the QoE of the workforce in heterogeneous vCS-mediated ubiquitous services, such as multi-modal immersion with latency-critical decision support, through e.g. AI/ML-driven network slicing \cite{Wu-Zhou}. Resource-aware interactions with real and synthesized applications executed in vCS are thereby fostered on top of   pillars of the Metaverse: Generative AI, collective intelligence, and transfer learning. Their combination strengthens automation efficiency pursued by Industry 4.0 and augment workers with anticipation and control capabilities for comfortable work \mbox{conditions  that Industry 5.0 targets.}  

\subsection{Generation of synthetic data to enrich and accelerate robotized workflows}
In global and decentralized data spaces, measured and synthetic data from geographically distributed locations with different modalities are structured using e.g.  Resource Description Framework (RDF)-links \cite{Heath}, as depicted in the bottom layer of \cref{metaverseconcept2}. Whereas measured data are obtained from sensing the states of  physical assets, such as a robot, synthetic data are generated using AI$\slash$ML techniques, such as a Generative Adversarial Network (GAN). A GAN employs the competition between a generator and  discriminator  to iteratively enhance its capability to synthesize  high quality and fidelity outputs, such as images, joint positions or torques, very close to and capturing the distribution of reference data, even under specific conditions \cite{Ku}. Hence, a conditional GAN (cGAN) is helpful to generate synthetic images of robot performing a collision avoidance, i.e., feasible paths \cite{Ma}. Such data can be used to train models, without the complexity associated with programming the physical robot or consuming energy during trial and error attempts to collect data. To this end, a loss function conditioned by the distribution of obstacles is used to construct  a latent space in which robot motions are collision-free \cite{Ando}. Furthermore,  the motion planning, i.e., generation of high-fidelity trajectories using the generator, is stripped down to connecting  configurations in the convex latent space \cite{Ando}.  Whereas the automated data generation for efficient and robust robot motion planning   aligns with Industry 4.0, its sustainable touch matches with goals of the Industry 5.0 concept. It is  worth to note that the generation of data is competitive in terms of efforts and costs. Small and medium sized enterprises that face scarce and limited data quality, can benefit from  generative AI$\slash$ML as functionality that the Metaverse combines with cross-domain simulation to strengthen their projection capabilities during co-innovation and speed up their workflows. In the Omniverse of NVIDIA, for instance, Omniverse Audio2Face offers an interface to generative AI$\slash$ML for the automated generation of facial data from an audio file \cite{audio2face}. Inferring  intentions of workers \cite{Wang2,Zhang} to achieve a symbiotic pHRC in Industry 4.0 and Industry 5.0 can benefit from such Metaverse-mediated data resources in an highly automated manner. Physically accurate motion data can also be collected by employing Move AI \cite{moveai} in the Omniverse to enhance safety and ergonomics in  human-centered robotics. Both objectives are prominent in  Industry 5.0 \cite{Zizic} and Industry 4.0 \cite{Baratta}. 

Another value offered by a decentralized data space  of the Metaverse in \cref{metaverseconcept2} to robotized applications is its pronounced  diversity and heterogeneity in terms of e.g. domain, location, and multi-modality. RDF allows for the storage of multi-modal scene graphs, including text, image, and video formats as well as underlying semantic similarities and spatial relations \cite{Li:Zheng}. Extensions of SPARQL \cite{Ali}, such as VGStore that builds on the grammar of SPARQL and the Python module \textit{pyparsing}, can be used to conveniently query multi-modal information on RDF-stored scene graphs \cite{Li:Zheng}. Handling multi-modality is pivotal to enrich pre-trained foundation models (e.g., BERT, CLIP, ChatGPT, etc.) \cite{Fei} provided in the collaboration space layer of the Metaverse in \cref{metaverseconcept2} for a twofold objective. First, introduce advances toward artificial general intelligence that attempt to mimic cognitive skills of humans \cite{Fei} into the closed control loops of robotized applications in  Industry 4.0 and Industry 5.0. Second, streamline the supervision and customization of robots as well as pHRC by  harnessing natural human capabilities, such \mbox{as spoken language (see \cref{natural}).}

\subsection{Collective knowledge creation for automation proficiency and symbiotic human-robot-collaboration}\label{collective}
In the second layer of the Metaverse, which is termed as knowledge space in \cref{metaverseconcept2}, learned knowledge graph representations are used to complete knowledge graphs, create and share new knowledge from distinct locations, accommodate uncertainties, and cope with unforeseen facts \cite{Daruna}. For this, confidence scores that encode and express
the level of belief in  relations is assigned to  pairs of  nodes of the knowledge graph. In this way,  uncertainties are taken into consideration \cite{Chen}. Markov logic networks, probabilistic soft logic, or embedding based approaches are possible directions to this end \cite{LiT}. In the context of robotized manufacturing, intentions of and  workpieces to be grasped and forwarded by the cobot to the carpenter in the case of co-worker vacancy  can be  inferred through graph completion \cite{Zhang}, fostering resilience in Industry 5.0. Inference skills are expected in the fifth predictability layer of the maturity index of Industry 4.0 \cite{Zeller}. Note also that the cobot assists and augments the carpenter through the gravity- and payload-compensated guidance of heavy workpieces ($\lambda = 1$). Therefore, the metabolic stress is \mbox{reduced, as desired in Industry 5.0.} 

\begin{figure}[t!]
	\centering
	\includegraphics[width=0.9\linewidth]{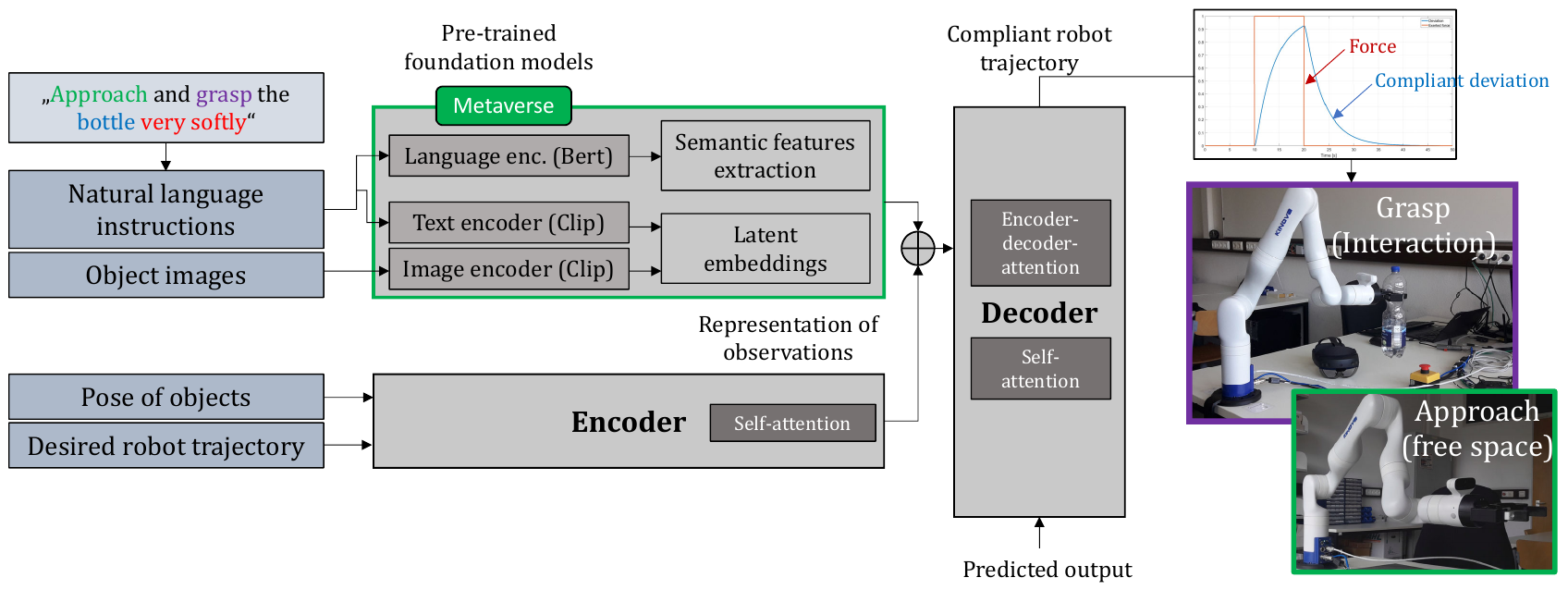}
	\caption{Architecture pursued in this work, which is based upon \cite{bucker2022latte}, for a natural (language-based) compliance control of  robots.}
	\label{fig:architecture}
\end{figure}

\subsection{Transfer learning for informed decision support and natural robot control} \label{natural}
The prediction and mining of knowledge in the second layer enrich the decision making process  with unforeseen insights in the third layer of the Metaverse called collaboration space in \cref{metaverseconcept2}. This happens for executives, engineers, and customers. cGAN-generated synthetic data can swiftly provide executives with potential use cases for robot design and extensions as well as robotized automation with customer-centered  economic, social, and ethical constraints at a high level of scalability. Reconfigurations of robotized workcells with enhanced collision avoidance to tap into new markets can be generated in a latent space \cite{Ando} using constrained instructions that executives formulate in a spoken language via e.g. BERT. The latent space is multidisciplinary engineered in the Metaverse through DT-based simulation. Feasibility and benefits of executive ideas are optimized. An advantage is the potentially fast and accurate transfer of  intentions of executives from a high level to different business units with less misunderstandings. Attention-based AI$\slash$ML-models can alert executives once  the performance of a robotized automation highly correlates with key terms spoken during a board meeting in real-time \cite{Ghojogh}. Transformer-based models can capture such correlations \cite{Choi}. Better-informed decision support is  irrespective of executive  locations. Executives collectively and simultaneously have access to  vCS as embodied avatars, in which case they can even show proximity to  customers by taking advantage of the Metaverse. Self-regulated blockchained synchronizations between the  real world and  Metaverse allow immersed customers to jointly modify the status of their orders (e.g., a  chair) while remaining aware of the  ecological footprint of the robotized automation used by the carpenter to complete their preferences in real-time.  The flexible proximity to executives as well as individualized and safe collaboration with workers, \mbox{e.g. the  carpenter,  elevates the customer QoE.} 

Instead of complex and lengthy robot programming, robotics  engineers can leverage natural language processing (NLP) to simplify the usage of the cobot \cite{bucker2022latte}. However, beyond \cite{bucker2022latte} that focuses on a geometric alteration  of the behavior of the robot to avoid collision, the vocabulary of the NLP-based instructions can be extended to adapt the compliant behavior of the robot in the joint space or at the Cartesian level. In this case, the carpenter and even novice collaborators, can control the cobot in a natural way.  Large scale pre-trained foundation models shared in the Metaverse, including Bert and Clip, can be used to semantically extract features from multiple modalities (see \cref{fig:architecture}). A connection is then established between the enriched vocabulary bank (e.g., "softly", "gently", "stiffly"), identified objects in the robot's vicinity, and a learnable admittance model. Multi-head self-attention mechanisms of the encoder of a transformer architecture is leveraged to this end. For instance, "softly" is highly correlated with a reduction of the desired stiffness of the admittance model. Compliance parameters are adapted to remain inside the stability area as in \cite{Ficuciello}. Similarly to \cite{bucker2022latte}, the decoder outputs modified trajectories using self-attention and cross-attention mechanisms. The modification is, however, force related. Specifically, the desired trajectory is modified according to a  second order mass, damper, and spring behavior of an admittance model between the sensed contact force and the deviation between the desired and compliant motion of the robot. It is worth noting that the compliant trajectory of the cobot is consistent to the deformation of geometrical trajectories developed by \cite{bucker2022latte} in the absence of  interaction forces. In fact, the compliant trajectory converged by definition of the admittance compliance model to the desired or geometrically modified one in the free-space (i.e., no physical interactions), as illustrated on the r.h.s of \cref{fig:architecture}. Therefore, the carpenter or any novices can naturally command a compliant behavior of the cobot without altering its free space motions. An advantage of  force-sensitiveness is that the compliant behavior is automatically activated once physical interactions are sensed without  any programming efforts. In this case, autonomous workpiece inference and collision-free grasping using the latent space ($\lambda = 0$),   kinesthetic manual guidance of the cobot with grasped payload and payload release ($\lambda \approx 0.5$) followed by kinesthetic manual guidance without payload  to the next customer order ($\lambda \approx 1$) becomes naturally feasible even for novices. However, the selection of a suitable robot might be challenging.

GAN-enabled Metaverse can accelerate transformation of insights gained from synthetic data into opportunities. GANs for discrete data with a generator trained using a policy gradient built upon the likelihood ratio from the discriminator facilitate the mapping of a robotics-related task objective gained from a knowledge inference (see \cref{collective}), such as ergonomic or energy-efficient grasping poses, to a distribution of robot designs \cite{Whitman:Hu}. This can be achieved even without a priori collected data and in a computationally efficient way \cite{Whitman:Hu}, \mbox{which highlights its resource-friendliness.}

\section{Conclusion}\label{conclusioon}
The paper has  introduced the concept of Metaverse, outlined some of its characteristics, and identified its potential advantages for robotized applications in different domains that build upon the Industry 4.0 and  Industry 5.0 visions. The generative, collective, and transferable role of AI$\slash$ML appears to be pivotal in terms of e.g. intrinsically safe inclusion of partners, enrichment of decision making, acceleration of the robotized workflow, and intuitive usage of robots. Opportunities to yield efficient, sustainable, resilient, and human-centered robotized applications have been highlighted from the perspective of workers, executives, and customers. Nevertheless, the Metaverse is still in its infancy. Standardization efforts for interoperability, security, as well as  privacy remain some of the plethora of open challenges. Furthermore, the development of pre-trained foundation models globally shared in the Metaverse and specialized through transfer learning to quickly solve e.g. the inverse kinematics or dynamics of {different classes of robots deserves further attention.} Future evaluations of the NLP-based control of robots are expected to reveal usability insights that benefit to a larger community because many use cases are possible. On the other hand, the Metaverse appears to be attractive for global and location-agnostic collaborations. Therefore, the support of pervasive and itinerant robotics-related co-innovation and product lifecycle management, regardless of the specific domain (e.g., education, manufacturing, elderly assistance, crisis management, healthcare, etc.), are potential future extensions.

\clearpage

\normalMode

\end{document}